\documentclass[10 pt, conference]{IEEEtran} 
\usepackage{geometry}
\IEEEoverridecommandlockouts                 

\usepackage{graphicx} 		
\usepackage{subfigure}		
\usepackage{amsmath} 		
\usepackage{amssymb}  		
\usepackage{algorithm}		
\usepackage{algpseudocode}
\usepackage{siunitx}		
\usepackage{subfigure} 		
\usepackage{balance}		
\usepackage{xcolor}			
\definecolor{dark-red}{rgb}{0.4,0.15,0.15}
\definecolor{dark-blue}{rgb}{0.2,0.2,0.6}
\definecolor{med-blue}{rgb}{0,0,0.5}
\definecolor{dark-green}{rgb}{0.0,0.4,0.0}

\newcommand{\tabref}[1]{Table~\ref{#1}}
\newcommand{\figref}[1]{Figure~\ref{#1}}
\renewcommand{\eqref}[1]{Equation~(\ref{#1})}
\renewcommand{\algref}[1]{Algorithm~(\ref{#1})}

\begin{document}
\title{Loop Closure Detection in Closed Environments}

\author{Nils Rottmann$^{1}$, Ralf Bruder$^{1}$, Achim Schweikard$^{1}$, Elmar Rueckert$^{1}$
 \thanks{$^{1}$Institute for Robotics and Cognitive Systems, 
	University of Luebeck,
        Ratzeburger Allee 160, 23562 Luebeck, Germany
         {\tt\small \{rottmann, bruder, schweikard, rueckert\}@rob.uni-luebeck.de}}%
} 


\maketitle



\begin{abstract}
Low cost robots, such as vacuum cleaners or lawn mowers employ simplistic and often random navigation policies. Although a large number of sophisticated mapping and planning approaches exist, they require additional sensors like LIDAR sensors, cameras or time of flight sensors. In this work, we propose a loop closure detection method based only on odometry data which can be generated using low-range or binary signal sensors together with simple wall following techniques. We show how to include the detected loop closing constraints into a pose graph formulation such that standard pose graph optimization techniques can be used for map estimation.\\ 
We evaluate our map estimate and loop closure approach using both, simulation and a real lawn mower in complex and realistic environments. Our results demonstrate that our approach generates accurate map estimates on the basis of odometry data only. We further show that our assumption about the discriminative nature of neighboring poses in the pose graph is solid, even under large odometry noise. These improved map estimates provide the basis for smart navigation policies in low cost robots and extends their abilities to goal-directed behavior like pick and place or complete coverage path planning in realistic environments. 
\end{abstract}

\section{Introduction}
\noindent During the last decade, research and development in the area of autonomous mobile robots have made significant progress. Nowadays robots such as vacuum cleaners or lawn mowers can be found in many households fulfilling their intended tasks. However, most of the robots employ thereby simplistic navigation strategies, such as a random walk, due to the lack of suitable maps and accurate sensors required for successful path planning. While most existing work for mapping and localization utilizes long-range sensors, such as LIDAR sensors, RGB-D cameras or time of flight sensors, robots for private households lack such sensor-richness. The reason for that is that such sensors are either too expensive, aiming for low acquisition and maintenance costs, or not suitable for outdoor environments, e.g. to reflections and direct sunlight. Moreover, simultaneous localization and mapping (SLAM) algorithms require certain amount of computational power which is often not available. However, intelligent navigation from low cost hardware is essential for mobile robots to enter our daily life. For example, autonomous lawn mowers employed with random walk policies are limited to simple environments, e.g. they can not enter small corridors. \\ 

\noindent In this paper, we propose a sophisticated mapping algorithm that can generate a map estimate of the environment based only on odometry data. Such data can be generated by following the wall or border line for an area of interest. Such wall following strategies can be executed using simple wall sensors, bumpers or signal wire sensors. The generated odometry data can be used for the proposed map estimation method. We will focus in this paper on how to achieve the most suitable map using only the recorded odometry data.\\

\noindent In order to reduce computational complexity, we first prune our problem using path segmentation. Thereby, we cluster the individual path points generated from the odometry data into straight line segments. Based on the pruned data set we generate a pose graph in which we search for loop closing constraints using shape comparison. We include these loop closing constraints into our pose graph formulation and optimize the graph using standard pose graph optimization techniques, such as the Levenberg-Marquardt algorithm. Finally, we introduce a measure for the mapping error based on the deviation of areas between the estimated map and the true shape of the environment.\\

\noindent We first evaluate our method in a complex simulation environment using standard motion models. Afterwards, we show how our method performs in a real robot, a Viking MI 422P, set up in a real garden environment. In order to represent real conditions as best as possible we learned the odometry parameters for the motion model by means of Log Likelihood estimation. Moreover, we show in simulations that our approach is robust even under a large amount of odometric drift, which is essential for successfully mapping outdoor environments.


\subsection{Contributions and Organization}

\noindent The contributions of the paper are two-fold. First, an efficient and simple method for loop closure detection using odometry only data is presented and second, a map evaluation scheme based on comparing two map areas, the estimated map and a groundtruth, is proposed. Both contributions are necessary to generate accurate map estimates with only low-range or binary sensors, which then enables low-cost robots to implement intelligent navigation and path planning strategies. These contributions are discussed in Section \ref{se:Methods}. In Section \ref{se:Results} we evaluate our approach in a realistic mowing scenario and in a challenging simulated apartment environment. We conclude in Section \ref{se:Conclusion}.

\subsection{Related Work}

\noindent Most of the existing work on SLAM \cite{montemerlo2002fastslam} or graph-based SLAM \cite{grisetti2010tutorial} relies on long-range sensors, such as LIDARs or cameras \cite{brenneke2003using}, \cite{civera2011towards}, \cite{konolige2008outdoor}, \cite{engelhard2011real}.  Most of these approaches are using sensor fusion and probabilistic reasoning, e.g. particle filter \cite{grisetti2007fast} or extended kalman filter \cite{bailey2006consistency}. However, there are some approaches which try to handle the SLAM problem using only sparse sensor data, e.g. \cite{beevers2006slam}, to avoid expensive sensing. Existing work for low-range sensors \cite{ozisik2016simultaneous}, \cite{choi2008line}, such as sonars or infrared sensors, requires linear features which are not necessarily present in outdoor environments. Indoor mapping with limited sensing using a wall following approach has been presented in \cite{zhang2010real}. However, this approach makes the assumption of an approximately rectilinear structure, which may be true for most indoor environments but not for outdoor applications. In contrast, our proposed approach is not restricted to such structural assumptions and can be used for either indoor or outdoor applications with arbitrary shapes. \\

\noindent Generating the path from odometry data leads to a pose-graph representation, which is often used for the SLAM problem \cite{olson2006fast}, \cite{latif2013robust} and has been first introduced by Lu and Milios 1997, \cite{lu1997globally}. Pose graph optimization has been addressed in several studies, e.g TORO \cite{grisetti2009nonlinear}, g2o \cite{kummerle2011g}, iSAM2 \cite{kaess2012isam2} and LAGO \cite{carlone2014fast}. Thereby, TORO is based on gradient descent and is an extension of Olson's algorithm \cite{olson2006fast}. It applies a tree-based parameterization to distribute residual errors across the graph that improves the performance. The "general graph optimization" framework, g2o, has been designed to perform the optimization of different least squares problems, which can be represented as a graph. It thereby relies on the Gauss-Newton method. The iSAM2 method applies Bayes trees using incremental variable re-ordering and fluid relinearization to solve sparse nonlinear incremental optimization problems. LAGO addresses the pose graph optimization problem by decoupling the orientation and translation. We use the simpler Levenberg-Marquardt algorithm \cite{grisetti2010tutorial}, which worked reasonably well for pose graph optimization. There, the function is approximated by its first order Taylor expansion around the current initial guess $\hat{\boldsymbol{p}}$ in order to then find the solution to the optimization problem iteratively. \\

\noindent In order to reduce computational complexity, we prune the data using path segmentation. Such a pruning can lead to a huge reduction of computational time at the price of a small increase in error \cite{latif2013go}.

\section{Methods}\label{se:Methods}
\noindent We start by introducing the standard pose graph formulation:\\
Let ${\boldsymbol{p} = \{\boldsymbol{p}_0,\dots,\boldsymbol{p}_N\}}$ be a set of $N+1$ poses representing the position and orientation of a mobile robot in a two dimensional space, hence $\boldsymbol{p}_i = [\boldsymbol{x}_i^{\top}, \varphi_i]^{\top}$. Here, $\boldsymbol{x}_i \in \mathbb{R}^2$ is the cartesian position of the robot and $\varphi_i \in [-\pi,\pi]$ the corresponding orientation as an euler angle with the integer $i = 0\mathit{:}N$. The relative measurement between two poses $i$ and $j$ is then given as
\begin{equation}
\label{eq:RelativeMeasurement}
\boldsymbol{\xi}_{ij} = \begin{bmatrix}
\boldsymbol{R}_i^{\top} (\boldsymbol{x}_j - \boldsymbol{x}_i) \\
\varphi_j - \varphi_i
\end{bmatrix} = \boldsymbol{p}_j  \circleddash \boldsymbol{p}_i,
\end{equation}
where $\boldsymbol{R}_i = \boldsymbol{R}_i(\varphi_i)$ is a planar rotation matrix and $\circleddash$ the pose compounding operator which has been introduced in \cite{lu1997globally}. These relative measurements are, in general, affected by noise. Thus, including a zero mean Gaussian noise ${\boldsymbol{\epsilon}_{ij} \sim \mathcal{N}(\boldsymbol{0}, \boldsymbol{P}_{ij})}$ leads to the noisy relative measurements
\begin{equation}
\label{eq:RelativeMeasurementsNoisy}
\hat{\boldsymbol{\xi}}_{ij} = \boldsymbol{\xi}_{ij} + \boldsymbol{\epsilon}_{ij} .
\end{equation}
In general, there are two different types of relative pose measurements: Odometric constraints and loop closing constraints. Here, the first constraints are generated by the wheel odometry of the differential drive robot. The second type of constraints are provided by the robot recognizing a match between actual measurements and past measurements by revisiting places. Subsection \ref{subse:LoopClosureDetection} shows how to efficiently identify and add loop closing constraints to the pose graph for odometry only data.\\

\begin{figure}[bt]
\centering
\includegraphics[width=0.48\textwidth]{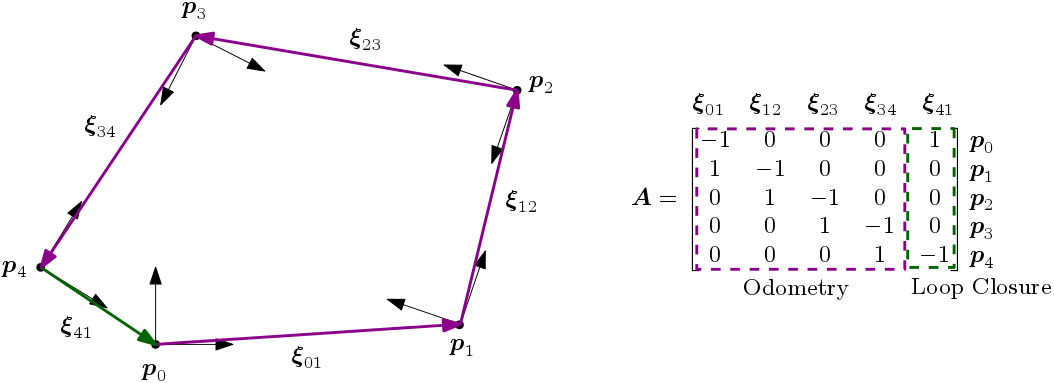}
\caption{Pose graph with five vertices connected with five edges. Four of the edges are odometric constraints and one is a loop closing constraint. On the right, the incidence matrix is shown divided into the parts containing the odometric constraints and the loop closing constraints.}
\label{fig:IncidenceMatrix}
\end{figure}

\noindent The pose graph is thereby represented as a directed graph ${\mathcal{G}(\mathcal{V},\mathcal{E})}$ with $N+1$ vertices and $N + M$ edges, where $N$ is the number of odometric constraints and $M$ the number of loop closing constraints. The connection between the vertices by the edges can be compactly written using an incident matrix $\boldsymbol{A}$. There, every column represents an edge connecting two vertices with each other. The row number thereby represents the vertex from which the edge starts, denoted by $-1$, and the vertex where the edge leads to, denoted by $1$. In \figref{fig:IncidenceMatrix} a pose graph is exemplarily shown in combination with the according incidence matrix $\boldsymbol{A}$.\\ 

\noindent The overall optimization problem is then to minimize the sum of weighted residual errors ${\boldsymbol{r}_{ij}(\boldsymbol{p})}$ in regard to the pose estimates $\boldsymbol{p}$
\begin{equation}
\label{eq:OptimizationProblemGeneral}
\min_{\boldsymbol{p}} \sum_{(i,j) \in \mathcal{E}} ||\boldsymbol{r}_{ij}(\boldsymbol{p})||^2_{\boldsymbol{P}_{ij}}
\end{equation}
with
\begin{equation}
\label{eq:ResidualError}
||\boldsymbol{r}_{ij}(\boldsymbol{p})||^2_{\boldsymbol{P}_{ij}} = [(\boldsymbol{p}_j  \circleddash \boldsymbol{p}_i) - \boldsymbol{\hat{\xi}}_{ij}]^{\top} \boldsymbol{P}_{ij}^{-1} [(\boldsymbol{p}_j  \circleddash \boldsymbol{p}_i) - \boldsymbol{\hat{\xi}}_{ij}] .
\end{equation}
The covariance matrix corresponding to the relative measurements $\boldsymbol{\hat{\xi}}_{ij}$ is thereby given as $\boldsymbol{P}_{ij}$. 

\subsection{Path Segmentation / Data Pruning}\label{subse:DataPruning}

\begin{algorithm}[tb]
\caption{DP Generation}
\label{alg:DPGeneration}
\begin{itemize}
\item \textbf{Parameters:} $L_{\text{min}}$, $e_{\text{max}}$
\item \textbf{Inputs:} $\boldsymbol{x}$	
\item \textbf{Outputs:} $DP$
\end{itemize}
\begin{algorithmic}[1]
\State $DP = [\boldsymbol{x}_0]$
\State $S = [\boldsymbol{x}_0]$
\If {new $\boldsymbol{x}$ available}
	\State $d = ||DP_{\text{end}} - \boldsymbol{x}||$
	\If {$d < L_{\text{min}}$}
		\State $S \leftarrow [S, \boldsymbol{x}]$
	\Else
		\State $S_{\text{tmp}} = [S, \boldsymbol{x}]$
		\State $e = \text{errorLineFit}(S_{\text{tmp}})$
		\If {$e < e_{\text{max}}$}
			\State $S \leftarrow S_{\text{tmp}}$
		\Else
			\State $DP \leftarrow [DP, S_{\text{end}}]$ 
			\State $S \leftarrow [S_{\text{end}}, \boldsymbol{x}]$
		\EndIf
	\EndIf
\EndIf
\end{algorithmic}
\end{algorithm}

\begin{figure}[tb]
\centering
\includegraphics[width=0.40\textwidth]{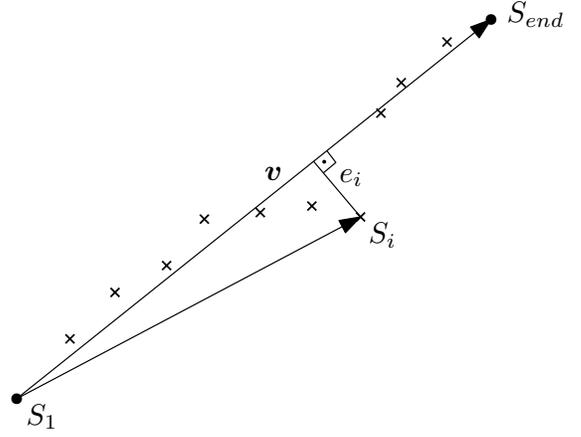}
\caption{To generate a new dominant point, the distance ${e_i \text{ with } j=2,\dots,|S|-1}$ is investigated. This distance represents the shortest distance between point $S_i$ and vector $\boldsymbol{v}$. More details are given in Subsection \ref{subse:DataPruning}.}
\label{fig:ShortestDistance}
\end{figure}

\noindent Path segmentation is used in order to cluster the individual path points retrieved from the odometry data into straight line segments. It is mostly used to reduce the complexity of the mapping problem, e.g. \cite{latif2013go} or \cite{zhang2010real}. The reduction of complexity will allow later improvements to the path by standard pose graph optimization techniques. Therefore, as shown in \figref{fig:ShortestDistance}, the errors between the individual odometry data points and a straight line segment are calculated and summed up to compare it with a certain threshold $e_{\text{max}}$. If the sum of the errors exceeds the threshold a new line segment with different orientation is generated. The path segmentation scheme is presented in \algref{alg:DPGeneration} and is inspired by \cite{zhang2010real}. We now shortly state the idea of the path segmentation approach.\\

\begin{figure}[tb]
\centering
\includegraphics[width=0.44\textwidth]{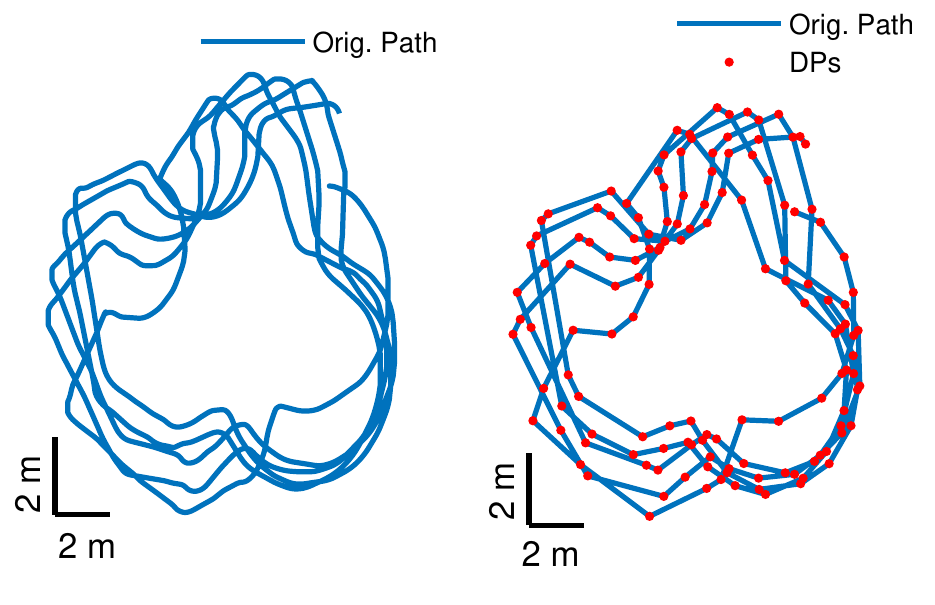}
\caption{Path Segmentation: In the left panel the original odometry data with 37687 data points is shown. In the right panel the path segmentation with 137 dominant points, marked as red dots, is presented.}
\label{fig:ExampleTraceSegmentation}
\end{figure}

\noindent Assume the position estimates based on the odometry are given as ${X = \{\boldsymbol{x}_0,\dots,\boldsymbol{x}_n\} }$. Here $n+1$ is the number of data points received from the odometry. First, a set of dominant points ${DP = \{ \boldsymbol{x}_0\} }$ and a temporarily subset ${S = \{\boldsymbol{x}_0\} }$ are initialized using the first position estimate $\boldsymbol{x}_0$. Hereby, the set of dominant points $DP$ represents the pruned data set and the temporarily subset $S$ contains the data points which are approximated by the current straight line segment.
We now successively include all odometry data points into the algorithm. Thereby, a new data point $\boldsymbol{x}$ is integrated into the temporary subset ${S \leftarrow \{S, \boldsymbol{x}\} }$. Afterwards, two conditions are checked:
\begin{equation}
\label{eq:MinimumLengthStraightLine}
L_{\text{min}} > ||DP_{\text{end}} - S_{\text{end}}||
\end{equation}
and
\begin{equation}
\label{eq:MaximumErrorStraightLine}
e_{\text{max}} >  \frac{1}{|S|-2} \sum_{i=2}^{|S|-1} e_i ,
\end{equation}
where $DP_{\text{end}}$ and $S_{\text{end}}$ are the last added items in the respective set. If both, \eqref{eq:MinimumLengthStraightLine} and \eqref{eq:MaximumErrorStraightLine}, are true, the temporary subset $S$ is not any longer a valid representation for the current straight line segment. Thus, the position $S_{\text{end}-1}$ is added to the set $DP$ and the temporary subset $S$ is set back to $S = \{S_{\text{end}}, \boldsymbol{x}\}$. Here, $e_i$ defines the shortest distance between the point $S_i$ and the vector $v = S_{\text{end}} - S_1$ and $|S|$ is the cardinality of the set $S$. In \figref{fig:ShortestDistance} the error calculation is exemplarily shown and in \figref{fig:ExampleTraceSegmentation} a resulting pruned data set is depicted compared to the original data set. The parameters $L_{\text{min}}$ and $e_{\text{max}}$ are problem specific and have to be tuned accordingly.\\

\begin{figure}[b]
\centering
\includegraphics[width=0.35\textwidth]{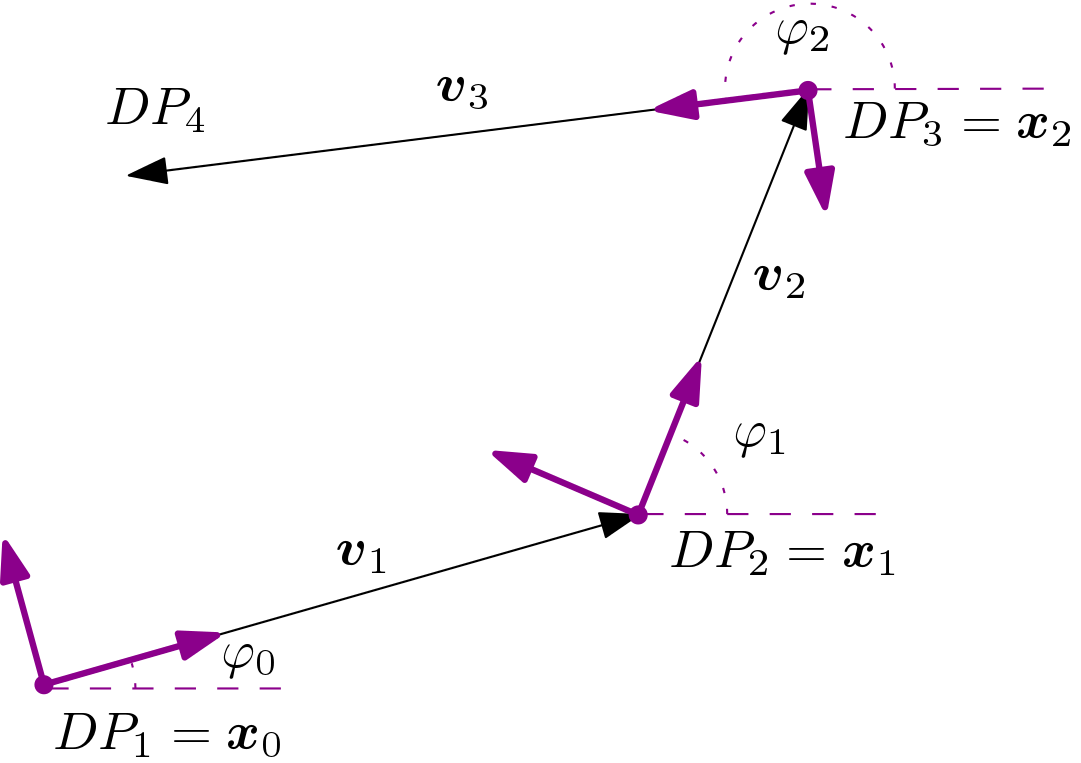}
\caption{The figure shows how the dominant points are transformed to a set of poses. Thereby, we start by with the first dominant point as the initial pose given by $\boldsymbol{p}_0 = [\boldsymbol{x}_0^{\top}, \varphi_0]^{\top}$ with $\boldsymbol{x}_0 = DP_1$ and $\varphi_0 = \text{atan2}(\boldsymbol{v}_{1,y},\boldsymbol{v}_{1,x})$, $\boldsymbol{v}_{1} = DP_2 - DP_1$.}
\label{fig:DP2Poses}
\end{figure}

\noindent Based on the pruned data set $DP$, the poses for the pose graph are generated as
\begin{equation}
\label{eq:PrunedPoses}
\boldsymbol{p} = \{ [DP_{1}^{\top}, \varphi_1]^{\top}, \dots, [DP_{|DP|-1}^{\top}, \varphi_{|DP|-1}]^{\top} \} ,
\end{equation}
where $|DP|$ is the cardinality of the set of dominant points and 
\begin{equation}
\varphi_i = \text{atan2}(\boldsymbol{v}_{i,y},\boldsymbol{v}_{i,x}) \quad \text{with} \quad \boldsymbol{v}_{i} = DP_{i+1} - DP_{i} .
\end{equation}
For the pose graph $N = |DP| - 2$ and the relative measurements $\boldsymbol{\hat{\xi}}$ can be calculated using \eqref{eq:RelativeMeasurement}. In \figref{fig:DP2Poses} an example for the generation of the pose graph based on the set of dominant points is shown. The part of the corresponding incidence matrix containing the odometric constraints can be then written as
\begin{equation}
A_{\text{Odometry}} =
\begin{bmatrix}
-1 & 0 & 0 & \dots  \\
1 & -1 &  & 		\\
0 & 1 & \ddots & \\
0 &  & \ddots & \\
\vdots & & & 
\end{bmatrix} \in \mathbb{R}^{N+1,N} .
\end{equation}

\subsection{Loop Closure Detection}\label{subse:LoopClosureDetection}

\noindent Only the pose data generated in Subsection \ref{subse:DataPruning} can be used to find loop closing constraints. Therefore, it is required for the robot to cycle several times along the boundary line of the area of interest. We then compare the shape of the path, searching for poses with similar shaped neighborhoods in order to find vertices which can be matched onto each other. These matched vertices can then be added as loop closing constraints to the pose graph.\\

\begin{figure}[bt]
\centering
\includegraphics[width=0.445\textwidth]{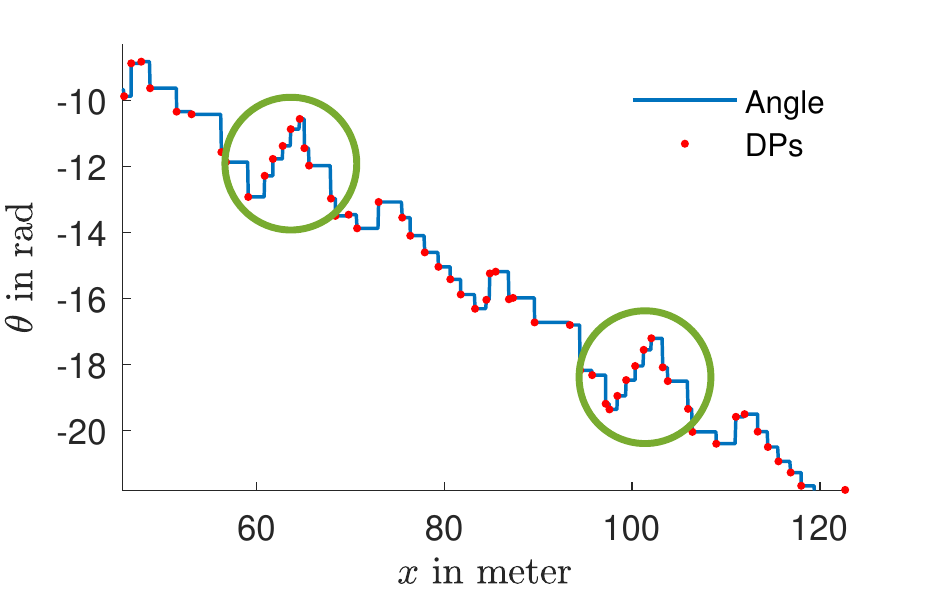}
\caption{Example for the piecewise linear orientation function $\theta(x)$. The green circled regions show similar path segments. The vertices of the pose graph are pictured as red dots.}
\label{fig:PiecewiseOrientationFunction}
\end{figure}

\noindent First, generate a piecewise linear function of the orientation in regard to the length of the path ${\theta = \theta(x)}$. The function is defined as
\begin{equation}
\label{eq:PiecewiseLinearFunctionOrientation}
\theta(x) = \phi_i \quad  \text{for}  \quad l_{i-1} \leq x < l_{i}, \quad i=0,1,\dots,N.
\end{equation}
The cumulated orientations $\phi_i$ and path lengths $l_i$ can be calculated as
\begin{equation}
\label{eq:cumulatedOrientationPathLength}
\begin{split}
\phi_{i} &= \phi_{i-1} + \Delta \phi_i \\
l_{i} &= l_{i-1} + ||\boldsymbol{v}_i||
\end{split}
\end{equation}
starting with ${\phi_0 = \varphi_0}$ and ${l_0 = 0}$ and going through all poses of the pruned graph. Here ${\boldsymbol{v}_i = \boldsymbol{x}_{i} - \boldsymbol{x}_{i-1}}$ and ${\Delta \phi_i = \varphi_{i} - \varphi_{i-1}}$ with $\boldsymbol{p}_i = [\boldsymbol{x}_i^{\top}, \varphi_i]^{\top}$. In \figref{fig:PiecewiseOrientationFunction} an example function based on the segmented data from \figref{fig:ExampleTraceSegmentation} is shown.\\

\begin{figure}[tb]
\centering
\includegraphics[width=0.49\textwidth]{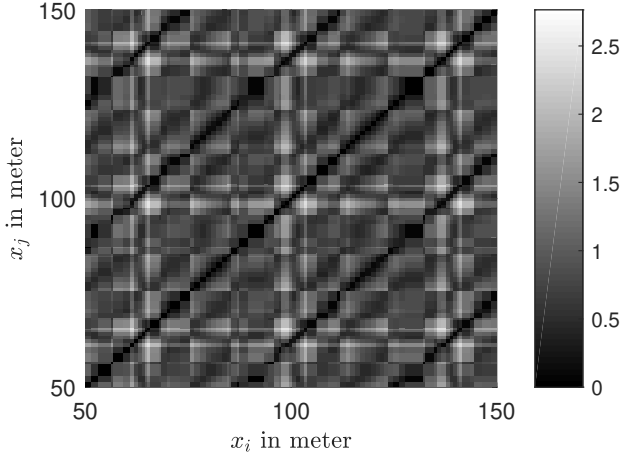}
\caption{Correlation error of the shapes of the neighborhood between the vertices of the pose graph. For better reading we plotted $\log(1 - \boldsymbol{C}_{ij})$ and only a section of the matrix. The variables $x_i$ and $x_j$ are representing the position $l$ of the vertices $i$ and $j$ in meter along the path.}
\label{fig:CorrelationError}
\end{figure}

\noindent Second, a comparison between the shapes of the neighborhood of the poses using the introduced piecewise orientation function ${\theta(x)}$ is done. Therefore, let $L_{\text{NH}}$ be the length to both sides of a pose which defines its neighborhood. A correlation error matrix $\boldsymbol{C}$ is then generated as follows:\\
Comparing the neighborhood of the pose $\boldsymbol{p}_i$ and $\boldsymbol{p}_j$, the orientation function $\theta(x)$ is grounded according to the orientations $\phi_i$, $\phi_j$ and lengths $l_i$, $l_j$ such that two neighborhood functions $\theta_i(x)$ and $\theta_j(x)$ are generated, e.g. ${\theta_i(x) = \theta(x+l_i) - \phi_i}$. Both functions are then evaluated at $m$ linearly distributed points from $-L_{\text{NH}}$ to $+L_{\text{NH}}$ which result into two vectors $\boldsymbol{\theta}_i$ and $\boldsymbol{\theta}_j$. The correlation error then add up to
\begin{equation}
\boldsymbol{C}_{ij} = \frac{1}{m} \sum_{k=1}^{m} ||\boldsymbol{\theta}_{i,k} - \boldsymbol{\theta}_{j,k}||^2 .
\end{equation}
In \figref{fig:CorrelationError} the correlation errors between the vertices are shown.\\

\noindent Third, we search for local minima which are below a threshold $c_{\text{min}}$ in the correlation error data. This leads to convenient pairs for loop closure $SP_k = \{\boldsymbol{p}_i,\boldsymbol{p}_j\}$ for $i \neq j$. In general, this procedure leads to a set of loop closing pairs $SP = [SP_1,SP_2,\dots,SP_M]$, where $M$ is the number of loop closing constraints found. The selection of an appropriate value for $L_{\text{NH}}$ is hereby crucial to find sufficiently accurate pairs for loop closure. In general, picking a larger value for $L_{\text{NH}}$ will lead to a more cautious selection and vice versa. Also, the threshold $c_{\text{min}}$ affects the loop closing detection significantly. Thereby, a huge neighborhood parameter $L_{\text{NH}}$ and large odometry errors should be compensated by choosing a large value for $c_{\text{min}}$ and vise versa.\\ 

\noindent The loop closing constraints can now be included into the pose graph representation by adding the relative measurements $\boldsymbol{\hat{\xi}}_{ij} = [0,0,0]^{\top}$ between the poses $i$ and $j$ for the loop closing pair ${SP_k = \{\boldsymbol{p}_i,\boldsymbol{p}_j\}}$. The incident matrix for the loop closing measurements can be built as
\begin{equation}
\label{eq:IncidenceMatrixLoopClosure}
\boldsymbol{A}_{\text{Loop Closing}} = \begin{bmatrix}
\boldsymbol{a}_1 & \dots & \boldsymbol{a}_M
\end{bmatrix} \in \mathbb{R}^{N+1,M}
\end{equation}
with the vector $\boldsymbol{a}_k$ representing the loop closing pair ${SP_k = \{\boldsymbol{p}_i,\boldsymbol{p}_j\}}$ by denoting $\boldsymbol{a}_{k,i} = -1$ and $\boldsymbol{a}_{k,j} = 1$ and the rest of the entries with zeros. The complete incidence matrix for the pose graph is then defined as
\begin{equation}
\label{eq:CompleteIncidenceMatrix}
\boldsymbol{A} = [\boldsymbol{A}_{\text{Odometry}}, \boldsymbol{A}_{\text{Loop Closing}}] \in \mathbb{R}^{N+1,N+M} .
\end{equation} 

\subsection{Pose Graph Optimization}\label{subse:MapOptimization}

\noindent Let ${\boldsymbol{e}_{ij}(\boldsymbol{p}_i,\boldsymbol{p}_j) = (\boldsymbol{p}_j  \circleddash \boldsymbol{p}_i) - \boldsymbol{\hat{\xi}}_{ij}}$ such that \eqref{eq:OptimizationProblemGeneral} can be rewritten as 
\begin{equation}
\label{eq:OptimizationProblemRewritten}
\min_{\boldsymbol{p}} \sum_{(i,j) \in \mathcal{E}} \boldsymbol{e}_{ij}^{\top} \boldsymbol{\Omega}_{ij} \boldsymbol{e}_{ij} ,
\end{equation}
where $\boldsymbol{\Omega}_{ij}$ is the information matrix for the relative measurement $\boldsymbol{\hat{\xi}}_{ij}$. To solve \eqref{eq:OptimizationProblemRewritten} the information matrices $\boldsymbol{\Omega}_{ij}$ have to be determined with $\boldsymbol{\Omega}_{ij} = \boldsymbol{P}_{ij}^{-1}$. For the odometric constraints we generate the covariance matrices as
\begin{equation}
\label{eq:OdometricCovariance}
\boldsymbol{P}_{\text{odometric},ij} = \text{diag}\left(
\begin{bmatrix}
\cos(\varphi_i) (\alpha_3 \delta_T + \alpha_4 \delta_R) \\
\sin(\varphi_i) (\alpha_3 \delta_T + \alpha_4 \delta_R) \\
\alpha_1 \delta_R + \alpha_2 \delta_T
\end{bmatrix} \right)
\end{equation}
on the basis of the odometry model presented in \cite{thrun2002probabilistic} and under the assumption that only one translation $\delta_T$ and one rotation $\delta_R$ occur. The parameters $\alpha_1,\dots,\alpha_4$ are robot specific and must be determined. For the loop closing constraints, the covariance matrices can be calculated using the correlation error from \eqref{fig:CorrelationError} and the parameters $\gamma_1$ and $\gamma_2$
\begin{equation}
\label{eq:LoopClosingCovariance}
\boldsymbol{P}_{\text{lc},ij} = \text{diag}\left(
\begin{bmatrix}
\gamma_1 & \gamma_1 & \gamma_2
\end{bmatrix} \right)
\boldsymbol{C}_{ij} .
\end{equation}
The parameters $\gamma_1$ and $\gamma_2$ can be used to improve the map estimate in complex environments. In our experiments all parameters were set to one. We then use the popular Levenberg-Marquardt algorithm as presented in \cite{grisetti2010tutorial} in order to solve the pose graph optimization problem from \eqref{eq:OptimizationProblemRewritten}.

\subsection{Map Generation and Evaluation}\label{SubSec:MapEvaluation}

\begin{figure}[b]
\centering
\includegraphics[width=0.48\textwidth]{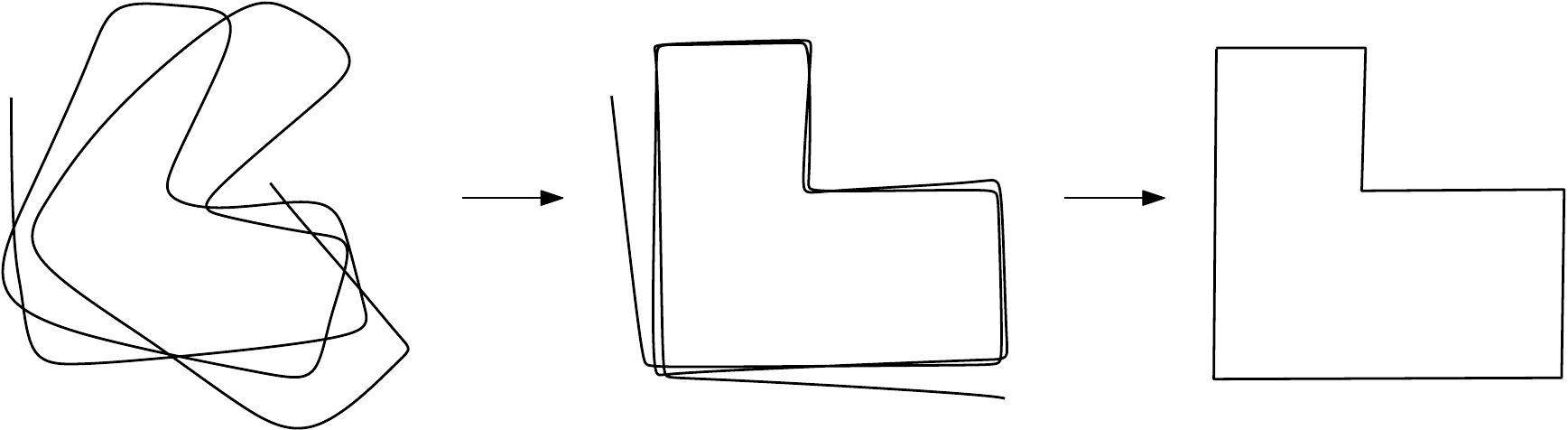}
\caption{The figures shows the steps we use for generating a polygon map of the boundary from the odometry data. The left panel shows the original odometry data, the mid panel the optimized pose graph and the right panel the closed polygon.}
\label{fig:MapGeneration}
\end{figure}

\noindent With Subsection \ref{subse:MapOptimization} we optimized our pose graph, for which an example can be seen in the mid panel of \figref{fig:MapGeneration}. Now, we shortly explain how to generate a closed trajectory from the optimized pose graph which can then be stored as a map of the environment. First, we cut off the prefix and the suffix of the pose graph, because these parts have not been optimized due to missing loop closing constraints. Thereby, the prefix is the part of the pose graph before the first loop closing constraint and the suffix the part of the pose graph after the last loop closing constraint. In order to generate a closed path, we investigate further the generated loop closing pairs. Here, we search for the first loop closing pair which represents a complete turn around the borderline. Setting the points of the this loop closing pair onto each other leads to a closed trajectory as presented in the right panel of \figref{fig:MapGeneration}. \\ 

\begin{figure}[bt]
\centering
\includegraphics[width=0.45\textwidth]{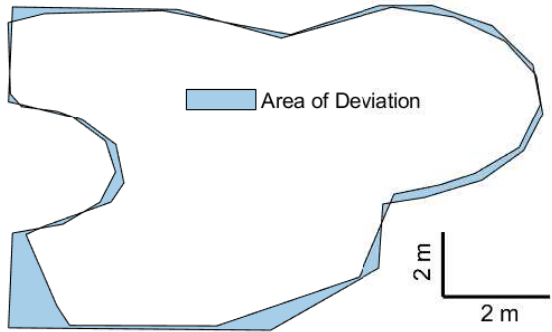}
\caption{Deviation of areas between an original map and the map estimate. The deviation is presented in light blue with $\Delta A = 9 \%$.}
\label{fig:DeviationOfAreas}
\end{figure}

\noindent To compare different map estimates with each other we calculate the deviation of area between the estimated map generated using the above algorithms and the true shape of the environment. Therefore, we assume that the true shape of the environment is given as a polygon defined by the points $X_{\text{true}}$. The idea is to determine a rotation matrix $\boldsymbol{R}$ and a translation vector $\boldsymbol{T}$ which transforms the set of points $X$, representing the closed map estimate, onto the set of points $X_{\text{true}}$
\begin{equation}
\label{eq:Transform}
\widehat{X} = \boldsymbol{R} \cdot X + \boldsymbol{T},
\end{equation}
such that 
\begin{equation}
\label{eq:DeltaA}
\Delta A = 1 - \frac{A_{\text{true}} \cap A_{\text{estimate}}}{A_{\text{true}} \cup A_{\text{estimate}}}
\end{equation}
is minimized. Here, $\Delta A$ represents the difference between the areas of the map estimate and the true shape of the environment. We use simple gradient descent in order to find a convenient rotation matrix and translation vector for \eqref{eq:Transform}. Since gradient descent requires a good initial guess in order to not get stuck into a local minimum, we use Horns method \cite{horn1988closed} to first find such initial guess. Therefore, we use the in Subsection \ref{subse:LoopClosureDetection} described method for loop closure detection in order to find pairs of points between the estimated map and the true shape of the environment which can be mapped onto each other. In \figref{fig:DeviationOfAreas} the deviation between the true shape of the environment and the estimated map is depicted in light blue.

\section{Results}\label{se:Results}
\begin{table}[b]
\centering
\caption{Measured Parameters for the odometry motion model (\cite{thrun2002probabilistic}).} 
\label{tab:ParamModels}
\begin{tabular}{c|c|c|c}
$\alpha_1$ & $\alpha_2$ & $\alpha_3$ &  $\alpha_4$  \\ 
\hline 
0.0849 & 0.0412 & 0.0316  & 0.0173 \\ 
\end{tabular}
\end{table}

\begin{figure*}[bt]
	\centering
	\subfigure[The courtyard of our Institute. We used the inner lawn area for testing the proposed mapping method.]{\includegraphics[width=0.31\textwidth]{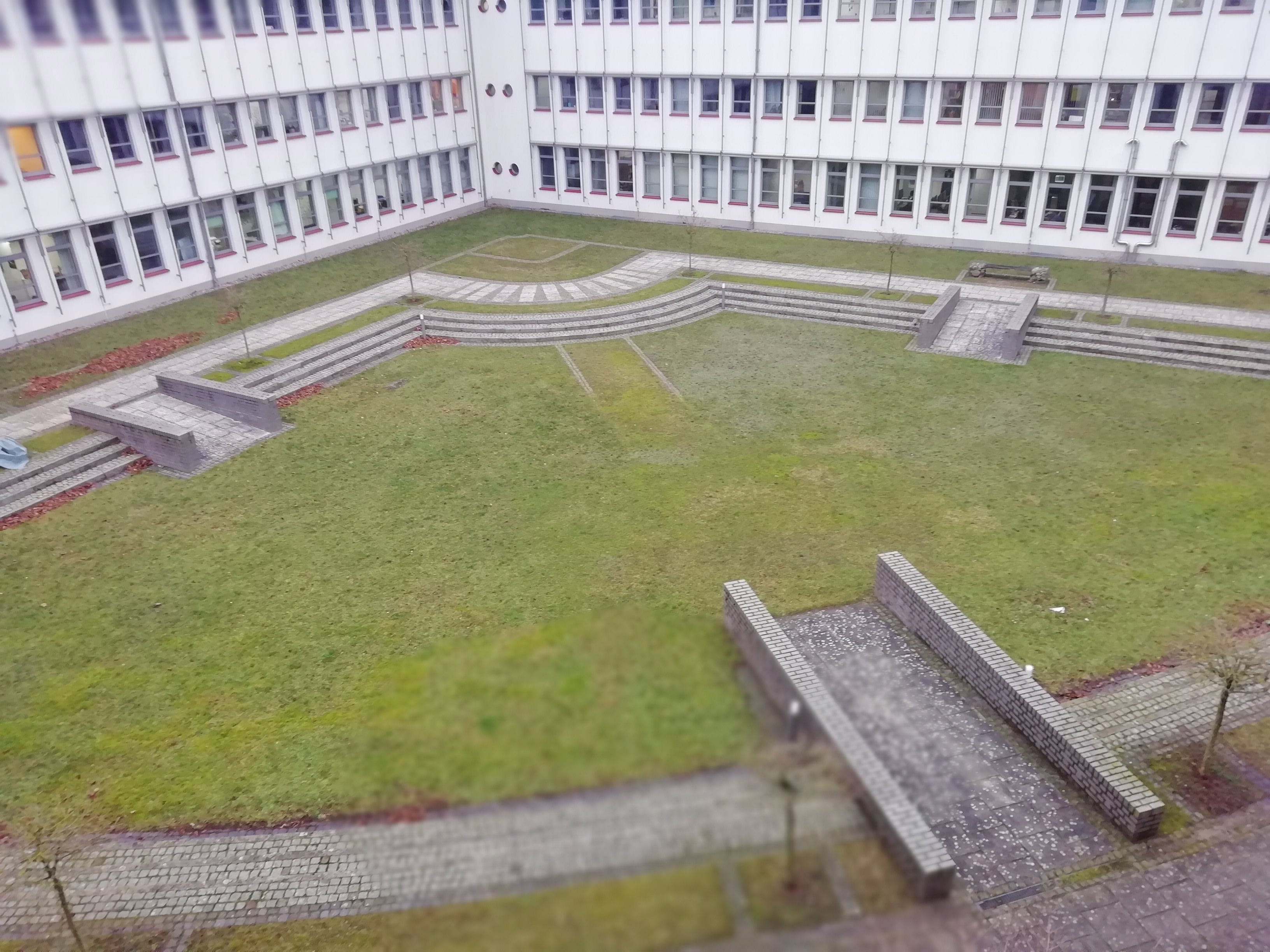}\label{subfig:RealCourtyard}}
	\hfill
	\subfigure[The left panel shows the estimated path of the robot generated from its wheel odometry and the right panel the estimated map and the true shape of the test environment.]{\includegraphics[width=0.62\textwidth]{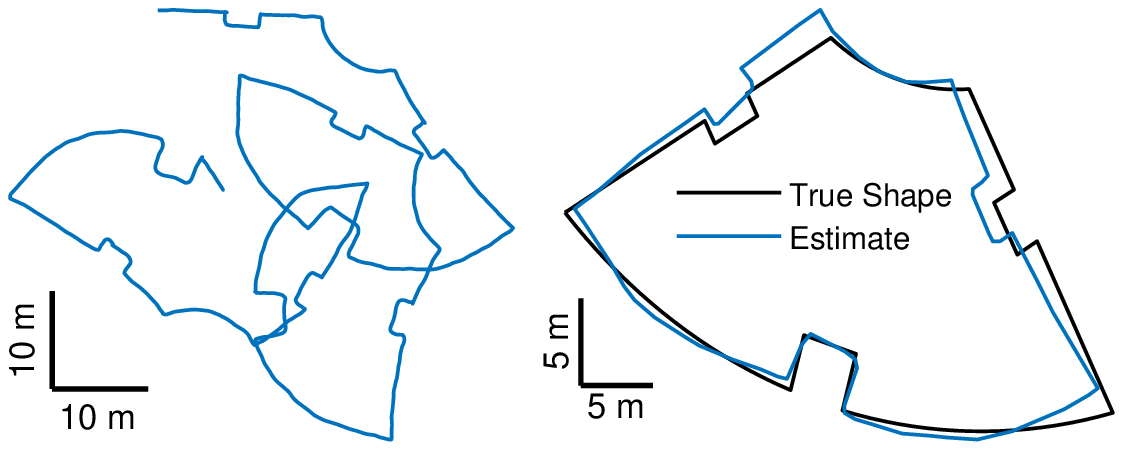}\label{subfig:RealCourtyardEstimate}}
		\caption{The real courtyard depicted in (a) and the collected odometry data together with the map estimate shown in (b).}
\end{figure*}

\noindent We tested the above proposed mapping method for odometry only data in simulations with challenging environments and on real data. For the real system, a Viking MI 422P, a purchasable autonomous lawn mower, has been used. For the simulation environment and for the estimation of the covariance matrices \eqref{eq:OdometricCovariance}, we used the odometry motion model presented in \cite{thrun2002probabilistic}. We calibrated the odometry model by tracking lawn mower movements using a visual tracking system (OptiTrack) and computed the parameters using maximum likelihood estimation. The calibrated parameters for the Viking MI 422P are presented in \tabref{tab:ParamModels}. For generating real data, we drove the lawn mower manually along the boundary line of the courtyard depicted in \figref{subfig:RealCourtyard}. Additionally, we simulated a differential drive robot with the specifications of the lawn mower in a challenging apartment environment. Moreover, we successively increased the odometry error in order to show that the proposed approach works even with large odometry errors and leads to sufficiently accurate results. As error measurement we used the deviation of areas between the true shape of the environment and the estimated map, as presented in Subsection \ref{SubSec:MapEvaluation}. Parameter specifications regarding the robot and the proposed method can be found in the Appendix, Section \ref{se:Appendix}.

\subsection{Apartment Environment - Simulation}

\begin{figure}[bt]
\centering
\includegraphics[width=0.26\textwidth]{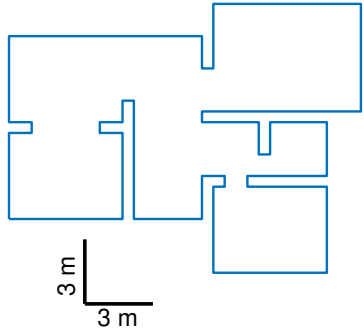}
\caption{The apartment envrionment used for the simulations.}
\label{fig:Apartment}
\end{figure}

\begin{figure}[bt]
\centering
\includegraphics[width=0.46\textwidth]{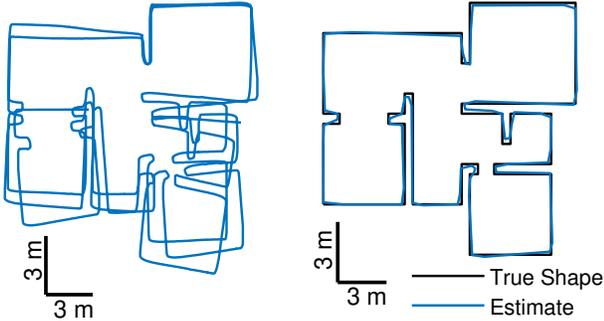}
\caption{The left panel shows the estimated path of the simulated robot generated using the introduced odometry model. The right panel shows the estimated map and the true shape of the apartment environment.}
\label{fig:ApartmentTrueParams}
\end{figure}

\noindent We used simulations in order to test our algorithm in a challenging environment, namely an apartment floor as depicted in \figref{fig:Apartment}. Since the apartment environment is complex we set the threshold $c_{\text{min}}$ to $1.0$. Simulating the differential drive robot using the in \tabref{tab:ParamModels} presented odometry parameters and a simple wall following algorithm results into the estimated path depicted in the left panel of \figref{fig:ApartmentTrueParams}. The resulting map estimate can be seen in the right panel together with the true shape of the apartment. The error between the true shape and estimated map is $\Delta A = 4.5\%$.\\

\begin{figure}[bt]
\centering
\includegraphics[width=0.47\textwidth]{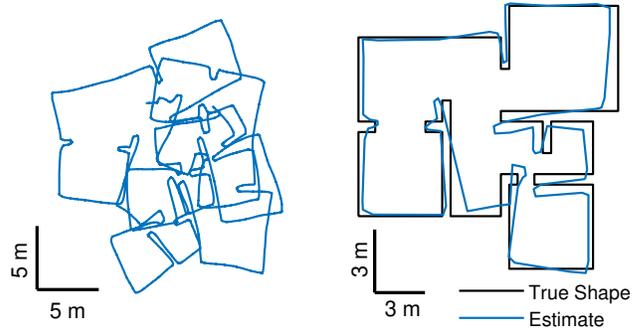}
\caption{The left panel shows the estimated path of the simulated robot generated using the introduced odometry model and harsher odometry parameter, $\alpha_i = 0.4$ for $i=1,\dots,4$. The right panel shows the estimated map and the true shape of the apartment environment.}
\label{fig:Apartment04}
\end{figure}

\noindent To simulate a drastic odometry drift, we set the odometry parameters to $\alpha_i = 0.4$ for $i=1,\dots,4$. This leads to an estimated path of the simulated robot as depicted in the left panel of \figref{fig:Apartment04}. The results of our map estimate is presented in the right panel of \figref{fig:Apartment04}. The mapping error is $\Delta A = 17.8\%$.\\
 
\noindent In order to show the robustness of our approach, we evaluated our method on the apartment environment for different values of $\alpha$ with $\alpha_i = \alpha$ for $i=1,\dots,4$. Therefore, we simulated the robot for every parameter setting $\alpha = \{0.1,0.2,\dots,0.5\}$ ten times and calculated the mean error $\mu_{\Delta A}$ and standard deviation $\sigma_{\Delta A}$, which can be seen in \tabref{tab:MeanSTD}. As expected, the mean $\mu_{\Delta A}$ rises monotonously with increasing $\alpha$ as well as the standard deviation $\sigma_{\Delta A}$.

\begin{table}[tb]
\centering
\caption{Mean values and standard deviation for different odometry parameters.}
\label{tab:MeanSTD}
\begin{tabular}{c|c|c|c|c|c}
$\alpha$ & $0.1$ & $0.2$ & $0.3$ & $0.4$ & $0.5$\\
\hline
$\mu_{\Delta A}$ & $7.91\%$ & $11.71\%$ & $14.01\%$ & $17.35\%$ & $27.30\%$ \\
$\sigma_{\Delta A}$ & $1.21\%$ & $2.09\%$ & $2.95\%$ & $3.79\%$ & $6.68\%$
\end{tabular}
\end{table}

\subsection{Courtyard Environment - Real Data}

\noindent For the real data set the parameter $c_{\text{min}}$ is set to $0.3$ to take into account the shape of the estimated path generated from the odometry data. This path can be seen in the left panel at \figref{subfig:RealCourtyardEstimate}. In the right panel, the map estimate generated using the proposed method is depicted together with the true shape of the courtyard environment. The error between both, according to the method presented in Subsection \ref{SubSec:MapEvaluation}, is $\Delta A = 11.87 \%$. The data for the courtyard has been retrieved from available CAD data.

\section{Conclusion}\label{se:Conclusion}

\noindent We have presented a method for map estimation based on odometry only data. This method is essential for cheap or small robots, such as vacuum cleaners or lawn mowers. Our method does not require any additional assumptions like for example a rectilinear structured environment \cite{zhang2010real} or linear features \cite{ozisik2016simultaneous}, \cite{choi2008line}. The required odometry data can be collected using a wall following scheme for which low range sensors or binary sensors are sufficient. Such sensors are widely used in actual purchasable household robots. Furthermore, we showed that our approach performs well in challenging environments, such as an apartment. Even  when simulating large odometry noise, our approach can generate an accurate map estimate.\\
The map estimate of the environment allows the robot to plan its intended task, such as cleaning the floor or mowing the lawn, instead of executing a random walk behavior. In a related work \cite{hess2014probabilistic}, probability distributions were used to update a dirt coverage map to keep the dirt level below a certain threshold. In another work \cite{bretl2013robust}, complete coverage is proven under the assumption of a bounded error. According to the first example a probability distribution encoded by a particle filter can be used to update a coverage map. This map allows then the execution of path planning strategies to avoid random walk behavior. This approach seems promising and will be further investigated.

\section{Appendix}\label{se:Appendix}
\noindent The velocity of the lawn mower driving along the boundary has been set to $\SI{0.3}{\metre\per\second}$. The odometry data has been sampled with a frequency of approximately $\SI{20}{Hz}$. The algorithmic parameters used for the map estimation can be found in \tabref{tab:Params}. Here the chosen parameters ${L_{\text{min}} = \SI{0.1}{\metre}}$, ${M = 100}$ and ${e_{\text{max}} = 0.001}$ for the path segmentation part reduce the complexity of the problem sufficiently while maintaining the path given by the odometry data. Moreover, the parameter for loop closure detection ${L_{\text{NH}} = \SI{30}{\metre}}$, since the used test environments have circumferences of $U_{\text{apartment}} = \SI{100}{\metre}$ and $U_{\text{courtyard}} = \SI{106.79}{\metre}$ respectively. Thus, slightly over $50\%$ of the length of the circumference is used for shape comparison. 
\begin{table}[tb]
\centering
\caption{Default Parameters for the proposed map estimation method.}
\label{tab:Params}
\begin{tabular}{c|c|c|c|c|c|c}
$L_{\text{min}}$  & $e_{\text{max}}$ & $L_{\text{NH}}$ & $M$ & $\gamma_1$ & $\gamma_2$ \\
\hline
$\SI{0.1}{\metre}$ & $0.001$ & $\SI{30}{\metre}$ & $100$ & $1.0$ & $1.0$ \\
\end{tabular}
\end{table}

\balance

\bibliography{papers}
\bibliographystyle{plain}


\end{document}